\documentclass[11pt]{article}

\usepackage{acl}
\usepackage{times}
\usepackage{latexsym}
\usepackage[T1]{fontenc}
\usepackage[utf8]{inputenc}
\usepackage{microtype}
\usepackage{inconsolata}
\usepackage{graphicx}
\usepackage{booktabs}
\usepackage{multirow}
\usepackage{amsmath}
\newcommand{\agent}[1]{\textit{#1}}

\title{When Faster Decisions Do Not Make Better Agents:\\An Empirical Study of Jev in Agent Workflows}

\author{
\textbf{Yuchen Ling}$^{1}$,
\textbf{Shengcheng Yu}$^{2,\dagger}$,
\textbf{Zhenyu Chen}$^{1}$,
\textbf{Chunrong Fang}$^{1,\dagger}$
\\
$^{1}$Nanjing University,
$^{2}$Technical University of Munich
\\
$^{1}$\texttt{yuchenling@smail.nju.edu.cn},
$^{2}$\texttt{shengcheng.yu@tum.de},\\
$^{1}$\texttt{\{zychen, fangchunrong\}@nju.edu.cn}
}

\begin{document}
\maketitle

\begin{abstract}

AI agents depend on repeated decisions to accomplish tasks, with large language models (LLMs) commonly serving as their core decision makers. The recent emergence of Jev offers an opportunity to make these decisions faster and at lower cost. However, it remains unclear whether such local savings can be translated into efficient and reliable task completion. In this paper, we investigate this efficiency transfer across five agents in two representative application settings, embodied tasks in ALFWorld and mobile app use in AndroidWorld. Across 4,500 task executions, we jointly evaluate task effectiveness, decision-level and task-level resource use, and execution behavior. Experimental results show that Jev integration substantially reduces decision cost and latency across all five agents, with task-level gains varying by workflow. Task success remains close to baseline levels on ALFWorld and declines more substantially in two mobile agents. Auxiliary reasoning and additional interactions absorb decision savings, while failed recovery and premature completion limit successful delivery. Mean cost per successful outcome increases in one agent even as spending per attempt falls. These findings highlight the importance of preserving progress, supporting recovery, and verifying completion when integrating efficient decision models into agent workflows. Their practical value should be assessed through successful task delivery, recovery behavior, and resources per delivered outcome.

\end{abstract}

\section{Introduction}

AI agents pursue goals through interaction with their environments, with LLMs supporting this process in different scenarios \citep{yang2024sweagent, wen2024autodroid, wang2024voyager}. Progress toward a goal depends on repeatedly interpreting observations and deciding what to do next \citep{yao2023react}. Some decisions require open-ended code or text generation, while others involve selecting among available tools, interface elements, or admissible commands \citep{wang2024voyager, wen2024autodroid, shridhar2021alfworld, yao2025taubench}. These bounded choices are repeated throughout a task, and make both action-selection quality and resource efficiency important to overall agent performance \citep{kapoor2025agents}.

The resources spent on these decisions should reflect their reasoning demands. On S1-Bench, \citet{zhang2026systemone} find that large reasoning models can overthink simple questions while still making errors, showing that extended reasoning does not consistently yield accurate, economical judgments. This finding motivates examining models designed for focused decisions. Jev, recently released by TypeSafe AI as a \emph{System One Model}, provides instruction-defined judgments through typed outputs, including probabilities over supplied choices \citep{almeida2026jev, typesafe2026primitives}. The company positions it as a fast, inexpensive decision model \citep{almeida2026jev}. Its choice interface offers an opportunity to reduce the cost and latency of repeated action selection in AI agents. Whether these local savings translate into efficient and reliable task completion remains unclear.

Answering this question requires following decisions through complete agent workflows. Actions shape subsequent observations and choices, so each decision affects the conditions for the next \citep{ross2011reduction}. An executable action can move an agent away from its goal, requiring further reasoning or recovery whose cost may offset local savings. Reflection extends these dependencies across attempts \citep{shinn2023reflexion}, and candidate construction shapes the choices available to the decision model \citep{yang2024sweagent}. Task efficiency therefore depends on auxiliary reasoning, interaction overhead, and the sequence of actions performed. Failed attempts also consume resources without delivering the requested outcome. Evaluating agent performance requires connecting decision-level savings to task success, overall resource use, and resources per successful outcome, supported by repeated evaluation and execution analysis \citep{kapoor2025agents, yao2025taubench}.

To provide this evidence, we conduct a comparative empirical study of AI agents with Jev-based action selection. The study covers five agent families in two representative application settings, embodied tasks in ALFWorld \citep{shridhar2021alfworld} and mobile app use in AndroidWorld \citep{rawles2025androidworld}. Within each family, we compare the original configuration with a Jev-integrated variant, using text-only observations to match Jev's supported inputs \citep{typesafe2026models}. Each configuration runs three times on the same tasks, yielding 4,500 task executions. We jointly evaluate task effectiveness, decision-level and task-level resource use, and execution behavior. Paired task-level analyses estimate uncertainty in the observed differences, while systematic trace coding characterizes failure and recovery patterns. Comparisons with the original configurations assess integration effects.

Our results show that Jev substantially reduces decision cost and latency across all five agents, with task-level gains varying by workflow. Task success remains close to baseline levels on ALFWorld and declines more substantially in two mobile agents. Mean task costs fall across all agents, while auxiliary reasoning and additional interactions limit overall speed gains. Failed recovery and premature completion further constrain successful delivery. Lower spending per attempt can therefore coexist with higher cost per successful outcome. These findings highlight the importance of preserving progress, supporting recovery, and verifying completion when integrating efficient decision models into agent workflows. Their practical value should be assessed through successful task delivery and the resources it requires. In summary, we make the following contributions:
\begin{itemize}
\setlength{\itemsep}{2pt}
\setlength{\parskip}{0pt}
\item We provide an empirical study of AI agents with Jev-based action selection across five agent families and two benchmarks, covering 4,500 task executions.
\item We characterize how decision-level savings translate into task cost, latency, and resources per successful outcome across workflows.
\item We systematically annotate execution traces and analyze recovery outcomes to identify failure and recovery patterns.
\end{itemize}
The experimental code is available at \url{https://anonymous.4open.science/r/jev-emp-5BF4/}.

\section{Background}

\textbf{Jev.} TypeSafe AI introduced Jev in September 2026 as a \emph{System One Model} for fast, structured decisions in software automation \citep{almeida2026jev}. Its API maps a shared state and instruction-defined questions to typed choices, scores, or binary probabilities. Questions are evaluated independently, and choice questions return probabilities over supplied candidates \citep{typesafe2026primitives}. Entailment-based classifiers and GLiClass support labels supplied at inference time \citep{yin-etal-2019-benchmarking, stepanov2025gliclass}. Jev's choice interface accepts candidate descriptions and instructions that define the decision to make. An agent can supply its goal, observations, and available actions as context for each selection. This connects the typed interface to repeated decisions within a workflow. Early Jev studies examine probabilistic coding of crash narratives \citep{rafe2026calibrated} and adoption across public software projects \citep{ling2026wild}. We examine how its structured decisions interact with reasoning, environment feedback, and repeated action selection during task execution.

\section{Related Work}

\textbf{Agent Action Selection.} LLM agents repeatedly map goals, observations, and interaction history to actions through reasoning and environment interaction \citep{yao2023react}. Their interfaces expose commands, UI operations, and tools across embodied, mobile, and software tasks \citep{wang2024voyager, wen2024autodroid, yang2024sweagent}. Tool use requires deciding which APIs to invoke, when to call them, and which arguments to supply \citep{schick2023toolformer}. A finite set of action types can still require open-ended arguments, so bounded selection and generation coexist within one agent. The surrounding workflow determines which candidates are available and how the selected action is executed. Feedback, reflection, and memory condition later decisions on earlier outcomes \citep{shinn2023reflexion, wang2024voyager, yang2026ossymphony}. We focus on the bounded-selection component, examining how a specialized decision model interacts with the workflow's reasoning, environment interaction, and recovery mechanisms.

\textbf{Efficient Decision Models and Routing.} Routing and cascades assign requests across models using difficulty, preferences, or confidence to balance cost and response quality \citep{ding2024hybrid, ong2025routellm, shen2025sater}. Agent systems extend model allocation to collaboration modes and roles through MasRouter \citep{yue2025masrouter}. EvoRoute uses experience and environment feedback for step-level backbone selection, evaluating performance, cost, and latency \citep{zhang2026evoroute}. These methods optimize which model a system invokes. We study specialized bounded action selection across five agent families, characterizing when decision-level savings transfer to task-level efficiency considering auxiliary reasoning, interaction length, recovery, and completion.

\textbf{Agent Evaluation and Sequential Reliability.} Benchmarks assess task outcomes across embodied, web, mobile, desktop, and software-engineering environments \citep{shridhar2021alfworld, zhou2024webarena, rawles2025androidworld, xie2024osworld, jimenez2024swebench}. ALFWorld and AndroidWorld provide the outcomes for our household and mobile evaluations. Resource-aware evaluation considers accuracy and expenditure, while repeated trials assess consistency \citep{kapoor2025agents, yao2025taubench}. Sequential errors can compound as earlier decisions alter the states encountered later \citep{ross2011reduction}. Feedback, reflection, and tree search support correction across attempts or alternative action paths \citep{shinn2023reflexion, zhou2024lats}. Failure analysis identifies difficulties in long-term reasoning, decision-making, and instruction following \citep{liu2024agentbench}. Trajectory diagnosis evaluates backtracking, observation reading, and self-verification in web agents \citep{ou2025agentdiagnose}. We connect decision-level resources to total expenditure and resources per delivered success, including failed attempts, and characterize recovery and completion errors through trace coding.

\section{Empirical Study Design}

Selecting among finite candidates matches Jev's stated focus on bounded judgments \citep{typesafe2026primitives}.
We therefore assess the benefits and limitations of Jev integration through task effectiveness, execution efficiency, inference cost, and consistency across runs, examining how decision-level savings translate into successful task delivery.

\subsection{Research Questions}

\paragraph{RQ1: Effectiveness.} How does integrating Jev into LLM-based agents affect task success relative to their original configurations?
\vspace{-0.5em}
\paragraph{RQ2: Efficiency Transfer.} When do decision-level efficiency gains transfer to the overall task-level efficiency?
\vspace{-0.5em}
\paragraph{RQ3: Failure and Recovery.} What failure patterns and recovery behaviors characterize agents with Jev-based action selection?

\subsection{Experimental Setup}

\textbf{Benchmarks.} We study action selection in two complementary interaction environments. AndroidWorld \citep{rawles2025androidworld} covers mobile application tasks, where agents act on changing UI elements and application states. We use text-based UI observations and one fixed parameter instance for each of its 116 task types. ALFWorld \citep{shridhar2021alfworld} covers household tasks through textual observations and admissible commands. We use its text environment and all 134 tasks in the \texttt{valid\_unseen} split. Together, these settings cover UI interaction and command-based control with language-based observations.

\textbf{Agent Configurations.} Since Jev accepts only text input currently, we select text-only agent configurations to match input modalities across variants. Table~\ref{tab:configurations} lists six configurations per benchmark, named \agent{Agent-B}, \agent{Agent-J}, and \agent{Agent-C} for Baseline, Jev, and LLM-Choice. AndroidWorld includes T3A \citep{rawles2025androidworld}, AutoDroid \citep{wen2024autodroid}, and Droidrun's FastAgent \citep{droidrun2026framework}, covering UI action generation, candidate-event selection, and tool invocation. Each has \textit{B} and \textit{J} variants, with FastAgent in text-only Direct mode and AutoDroid's app-specific memory disabled in both. ALFWorld includes ReAct \citep{yao2023react} and Reflexion \citep{shinn2023reflexion}, which support within-trial reasoning and cross-trial reflection, respectively. Each has \textit{B}, \textit{J}, and \textit{C} variants. The \textit{C} control is needed because Jev integration adds admissible-command access as well as changing the decision model. It uses an LLM with Jev's candidate list and selection instructions. \textit{B}--\textit{C} comparisons therefore assess the candidate-selection interface, while \textit{C}--\textit{J} comparisons assess the decision model under that interface.

\begin{table}[!htbp]
\centering
\small
\setlength{\tabcolsep}{2pt}
\renewcommand{\arraystretch}{1.12}
\begin{tabular*}{\columnwidth}{@{\extracolsep{\fill}}lll@{}}
\toprule
\textbf{Benchmark} & \textbf{Agent} & \textbf{Variants} \\
\midrule
\multirow{3}{*}{\shortstack[l]{AndroidWorld\\\scriptsize\citep{rawles2025androidworld}}}
 & T3A {\scriptsize\citep{rawles2025androidworld}} & \textit{B, J} \\
 & AutoDroid {\scriptsize\citep{wen2024autodroid}} & \textit{B, J} \\
 & FastAgent {\scriptsize\citep{droidrun2026framework}} & \textit{B, J} \\
\midrule
\multirow{2}{*}{\shortstack[l]{ALFWorld\\\scriptsize\citep{shridhar2021alfworld}}}
 & ReAct {\scriptsize\citep{yao2023react}} & \textit{B, J, C} \\
 & Reflexion {\scriptsize\citep{shinn2023reflexion}} & \textit{B, J, C} \\
\bottomrule
\end{tabular*}
\caption{Six configurations per benchmark. The suffix B denotes the baseline with its original LLM decision strategy. J uses Jev for action selection. C denotes LLM-Choice, which uses an LLM with the same candidates and selection instructions as Jev.}
\label{tab:configurations}
\end{table}

\textbf{Jev Integration.} We replace action selection with one logical Jev request per decision step, retaining the surrounding observation, history, and execution mechanisms. In AndroidWorld, Jev selects UI actions and targets for T3A, candidate events for AutoDroid, and tools with associated arguments for FastAgent. In ALFWorld, it selects an admissible command or THINK, which invokes LLM reasoning. Auxiliary LLMs supply open-ended text and summaries, including Reflexion's reflections. The comparisons include two interface differences: ALFWorld baselines generate commands without candidate lists, and \agent{FastAgent-J} executes one tool per step while \agent{FastAgent-B} can emit multiple calls.

\textbf{Execution Protocol.} We run each configuration three times on the same task suite with seed~0 to measure variability on fixed instances. All LLM decisions and auxiliary generation use \texttt{gpt-5.6-luna}, and Jev decisions use \texttt{jev-1.13.0}. Environment state and memory are reset between tasks and repetitions, with Reflexion memory retained only across trials of the same task. Budgets are matched within each agent family. AndroidWorld allows 600 seconds of agent execution and $\lceil 10c\rceil$ steps per task, where $c$ is task complexity. ALFWorld allows ReAct 49 steps and 600 seconds per task. Reflexion allows up to ten trials of 49 steps within 6,000 seconds, stopping upon success. Cross-family comparisons thus reflect different budgets as well as agent designs.

\subsection{Evaluation and Metrics}

\textbf{Repeated Runs and Aggregation.} To capture consistency and variability in agent execution, we evaluate each agent configuration three times on the same tasks, resetting environment state and agent memory between runs. We retain all task--run records and report the mean and sample standard deviation of each metric across runs. Paired analyses preserve task correspondence. We report \textit{J} minus control differences with 95\% task-cluster bootstrap intervals, resampling tasks and retaining their three runs together in 20,000 replicates. See Appendix~\ref{sec:cluster-bootstrap} for details.

\textbf{Effectiveness.} Our primary metric is task success rate, the proportion of tasks completed successfully according to the benchmark's official evaluator. For AndroidWorld, we additionally report the official task score to capture partial completion.

\textbf{Efficiency and Cost.} We measure token usage, inference cost, and elapsed time at both task and decision levels. Task-level token usage and cost include all model calls, including auxiliary generation and reflection, while decision-level metrics cover only action-selection calls. Inference costs are calculated using fixed model-specific input and output token prices. We report each metric over all tasks and over successful tasks only, with the latter including only decisions from successful executions. Task-level means divide the corresponding totals by the number of tasks, and decision-level means divide them by the number of action-selection calls within each scope. We also report cost and time per delivered success by dividing the corresponding totals across all attempts, including failures, by the number of successful executions.

\section{Results and Analysis}

\subsection{RQ1: Effectiveness}

\begin{figure}[!htbp]
\centering
\includegraphics[width=\columnwidth]{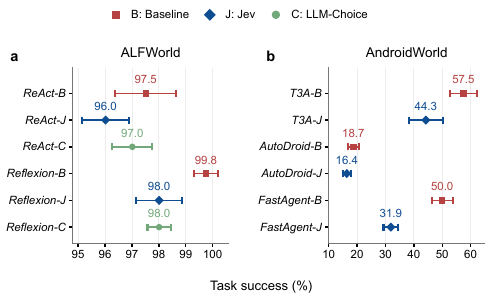}
\caption{Task success rates on ALFWorld (a) and AndroidWorld (b), shown as means $\pm$ one sample SD across three runs. Horizontal axis ranges differ.}
\label{fig:final-effectiveness}
\end{figure}

\textbf{An uneven effectiveness gap.} On ALFWorld, \agent{ReAct-J} and \agent{Reflexion-J} achieve 96.02\% and 98.01\% success, trailing their original configurations by 1.49 and 1.74 percentage points (pp), respectively (Figure~\ref{fig:final-effectiveness}). With the shared candidate interface in C, these gaps narrow to 1.00 and 0.00 pp. On AndroidWorld, Jev trails T3A, AutoDroid, and FastAgent by 13.22, 2.30, and 18.10 pp, respectively, with both AutoDroid variants below 20\% success. Comparisons with original configurations capture system-level integration effects, including an interface change on ALFWorld. AndroidWorld agents already use predefined action lists. Jev thus retains nearly all success under ALFWorld's shared interface but incurs larger losses in T3A and FastAgent. The 95\% task-cluster bootstrap intervals for J--B success differences are entirely negative for T3A and FastAgent and include zero for the other agents and both ALFWorld J--C comparisons (Table~\ref{tab:bootstrap}). Appendix~\ref{sec:supp-effectiveness} reports supplementary results.

\begin{figure}[!htbp]
\centering
\includegraphics[width=\columnwidth]{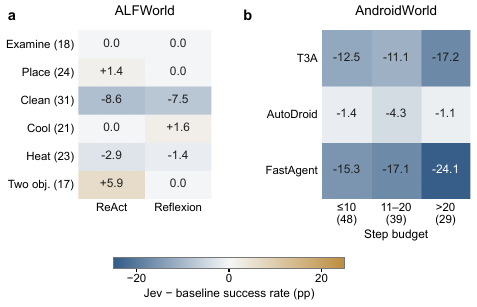}
\caption{Jev--baseline success-rate differences (pp) by ALFWorld task category (a) and AndroidWorld predefined step budget (b), pooled over three runs.}
\label{fig:final-conditions}
\end{figure}

\textbf{Task difficulty extends beyond sequence length.} Within these workflows, success varies with task demands (Figure~\ref{fig:final-conditions}). Both Jev families reach 100\% success on ALFWorld's examine, place, and cool tasks, and Jev improves ReAct's success on two-object tasks. Cleaning produces deficits of about 8.6 pp and 7.5 pp, with a similar weakness in LLM-Choice. On AndroidWorld, T3A and FastAgent lose 12.5 and 15.3 pp even in the shortest-budget stratum, with larger deficits in the longest-budget stratum. Jev can sustain successful sequences with multiple subgoals, yet also lose effectiveness on short tasks. The longer-task deficits indicate an additional challenge, while the short-task failures direct attention to the judgments required within individual states. This pattern makes the content of a decision, including its target, prerequisites, and completion conditions, central to explaining the observed differences.

\textbf{Errors become failures through termination or failed recovery.} Trajectories connect these task differences to goal interpretation and correction. In the Wi-Fi task, Jev selects an ambiguously labeled switch and declares completion with the goal unmet. This terminal judgment ends the opportunity for correction. In ALFWorld, selecting a handtowel for a cloth leads the workflow to continue reasoning on the wrong object. The initial error is amplified as the agent repeatedly preserves the mistaken premise. These cases show how a decisive stopping error can end a short task, while an earlier selection error can develop into sustained failure through unsuccessful correction. Reflexion demonstrates recovery's contribution at the task level, as Jev's first-trial success advantage becomes a final deficit after further trials (Table~\ref{tab:reflexion-recovery}). Its selection and auxiliary LLM reflection jointly determine this outcome. Effective decision making consequently requires both selecting a goal-appropriate action and revising previous choices when feedback reveals a mismatch.

\textbf{Jev suits clearly specified decision roles.} Jev is most promising when the decision input makes the goal, current state, and action prerequisites explicit, so selecting an action maps an understood state to a suitable next operation. Promising applications include workflow routing, API or tool selection, and routine software operations with identifiable targets and observable completion conditions. Workflows involving ambiguous goals, incomplete observations, or repeated sub-goal revision place greater demands on interpretation and recovery, the capabilities implicated in the observed mobile deficits. The effectiveness trade-off is strong execution within well-specified states versus adaptability when the agent must revise its understanding of the task. The strongest candidates for Jev have explicit conditions for choosing, checking, and completing an action.

\subsection{RQ2: Efficiency Transfer}

\begin{figure*}[!htbp]
\centering
\includegraphics[width=\textwidth]{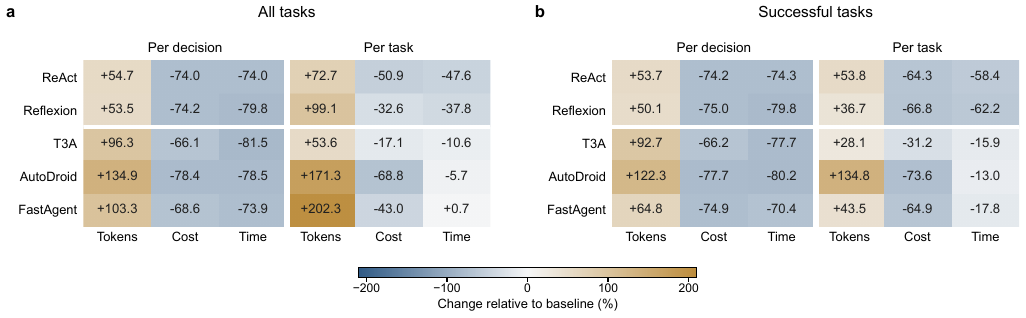}
\caption{Jev's resource changes relative to B (\%), from run-averaged means. Panels include all tasks (a) or each configuration's successful tasks (b). Per-decision metrics cover action-selection calls. The first two rows are ALFWorld and the remaining rows AndroidWorld. Blue indicates decreases and gold increases.}
\label{fig:final-resources}
\end{figure*}

\textbf{Faster and cheaper action selection.} Across all five families, Jev reduces mean decision cost by 66.1\% -- 78.4\% and time by 73.9\% -- 81.5\% (Figure~\ref{fig:final-resources}a). Decision times fall from 4.13 -- 8.13 to 1.04 -- 1.75 seconds (Appendix~\ref{sec:supp-resources}), with both advantages persisting among successful tasks (Figure~\ref{fig:final-resources}b). Tokens per decision increase by 53.5\% --134.9\%, as selection consumes observations, choices, and history. Jev's lower input price and zero output charge offset this increase.

\textbf{Cost savings transfer more consistently than speed.} Task-level gains are smaller and vary across workflows (Figure~\ref{fig:final-resources}a). On ALFWorld, ReAct and Reflexion reduce cost by 50.9\% and 32.6\%, and time by 47.6\% and 37.8\%, respectively. Against C, cost falls by 56.6\% and 63.9\%, with approximately 56.6\% shorter task times (Appendix~\ref{sec:supp-resources}). On AndroidWorld, T3A, AutoDroid, and FastAgent reduce cost by 17.1\%, 68.8\%, and 43.0\%. Time falls by 10.6\% and 5.7\% for T3A and AutoDroid and rises by 0.7\% for FastAgent. All families use more tokens per task. Paired 95\% task-cluster bootstrap intervals support lower costs in all J--B comparisons except Reflexion, whose interval includes zero. Time intervals include zero for AutoDroid and FastAgent (Table~\ref{tab:bootstrap}). Cost savings thus transfer more consistently across workflows. These task-level comparisons capture the combined effects of action selection, auxiliary calls, and environment interaction under each integration.

\begin{figure}[!htbp]
\centering
\includegraphics[width=\columnwidth]{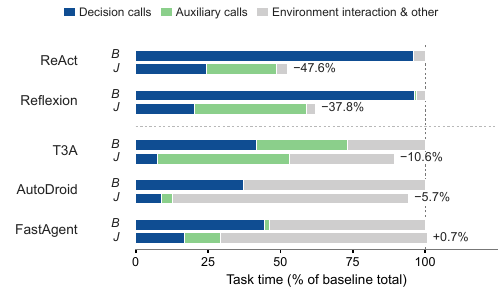}
\caption{Mean task-time components over three runs, normalized to each family's baseline (B = 100\%). J labels show total-time changes. Gray denotes environment interaction and other overhead. The divider separates ALFWorld (top) and AndroidWorld (bottom).}
\label{fig:time-decomposition}
\end{figure}

\textbf{Work outside selection absorbs the savings.} Transfer depends on the initial share of action selection and additional workflow overhead (Figure~\ref{fig:time-decomposition}). Baseline ReAct spends 68.31 of 71.17 seconds selecting actions. Jev reduces selection to 17.25 seconds, retaining substantial savings after adding 17.23 seconds of auxiliary reasoning. AutoDroid spends 164.43 of 261.20 seconds outside model calls. This component rises to 213.37 seconds with Jev, absorbing much of the selection saving. T3A's auxiliary generation grows from 67.47 to 96.99 seconds and accounts for 81.9\% of Jev's inference expenditure, limiting cost savings. FastAgent makes 44.1\% more decisions and spends longer on auxiliary calls and non-model execution. Its one-tool-per-step configuration combines cheaper decisions with longer interaction sequences. Task gains depend on both the share accelerated and the total work performed.

\textbf{Completion determines the value of savings.} Every family is faster and cheaper on its own successful-task subset, which differs between configurations. On matched task--run pairs completed by both, ALFWorld retains 64.8--66.2\% lower cost and 59.2--60.8\% shorter execution (Table~\ref{tab:resource-delivery}a). AndroidWorld retains cost savings, while FastAgent's time saving narrows to 5.5\% and T3A becomes 1.1\% slower. Paired cost intervals are below zero for all families, while time intervals include zero for T3A and FastAgent (Table~\ref{tab:bootstrap}). Failures add to delivery costs. In \agent{Reflexion-J}, 2.0\% of executions consume 53.3\% of inference expenditure through unsuccessful recovery (Table~\ref{tab:resource-delivery}b). Across all attempts, cost per delivered success falls in four families and rises by 7.7\% for T3A, from 68.70 to 73.97 mUSD (Table~\ref{tab:resource-delivery}c). This increase has a 95\% interval spanning zero. At the observed means, lower completion offsets T3A's 17.1\% cost saving per attempt. FastAgent's nearly unchanged task time and lower completion raise time per delivered success from 239.4 to 377.8 seconds. T3A also takes longer per success, with positive difference intervals for both agents. Completion and failed execution therefore shape the final resource benefit.

\begin{table}[!htbp]
\centering\footnotesize
\setlength{\tabcolsep}{1.5pt}
\renewcommand{\arraystretch}{1.08}
\begin{tabular*}{\columnwidth}{@{\extracolsep{\fill}}lrrrrr@{}}
\toprule
 & ReAct & Reflexion & T3A & AutoDroid & FastAgent \\
\midrule
\multicolumn{6}{@{}l}{\textbf{(a) Matched successes}} \\
Pairs & 378 & 393 & 145 & 50 & 84 \\
Cost (\%) & -64.8 & -66.2 & -11.3 & -75.3 & -50.8 \\
Time (\%) & -59.2 & -60.8 & +1.1 & -18.6 & -5.5 \\
\midrule
\multicolumn{6}{@{}l}{\textbf{(b) Cost spent on failures (\%)}} \\
B & 8.7 & 3.5 & 55.0 & 92.8 & 56.0 \\
J & 34.6 & 53.3 & 71.2 & 94.7 & 82.7 \\
\midrule
\multicolumn{6}{@{}l}{\textbf{(c) Resources per delivered success}} \\
Cost B & 7.93 & 9.00 & 68.70 & 27.93 & 21.96 \\
Cost J & 3.96 & 6.18 & 73.97 & 9.93 & 19.62 \\
Time B & 73.0 & 99.8 & 370.6 & 1398.4 & 239.4 \\
Time J & 38.9 & 63.2 & 430.2 & 1503.5 & 377.8 \\
\bottomrule
\end{tabular*}
\caption{Efficiency of successful delivery. (a) Changes from B to J on matched successes. (b) Failure share of inference cost. (c) Cost (mUSD) and time (s) per success, including failed attempts. Paired 95\% intervals appear in Table~\ref{tab:bootstrap}.}
\label{tab:resource-delivery}
\label{tab:matched-delivery}
\end{table}

\textbf{Efficiency transfer depends on workflow overhead and completion.} Decision savings transfer most directly when selection dominates resource use, auxiliary reasoning and additional interactions remain limited, and task success is preserved. ReAct meets these conditions and lowers both cost and time per delivered success. Auxiliary work, extra interactions, and lower completion weaken gains in T3A and FastAgent (Figure~\ref{fig:time-decomposition}, Table~\ref{tab:resource-delivery}). Tool routing and software-service orchestration with lightweight operations and little auxiliary generation are promising settings. Slow external operations limit time savings. Practical value depends on workflow resource use, task completion, and the cost of unsuccessful execution.

\subsection{RQ3: Failure and Recovery}

\begin{figure}[!htbp]
\centering
\includegraphics[width=\columnwidth]{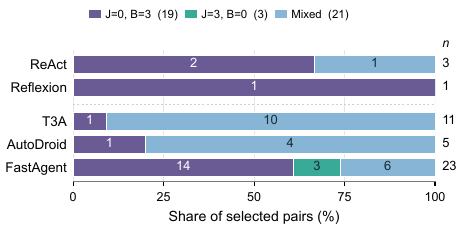}
\caption{Composition of selected outcome groups. Bar segments show within-family proportions and exact counts, with totals at right. $J$ and $B$ count successes over three runs. Mixed means both succeed once or twice. The dashed line separates the benchmarks.}
\label{fig:case-summary}
\end{figure}

\begin{figure*}[!htbp]
\centering
\includegraphics[width=\textwidth]{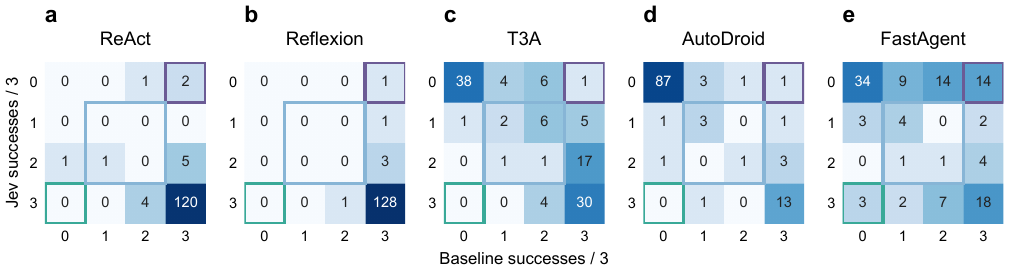}
\caption{Task counts by Jev and baseline successes over three runs, on a shared logarithmic color scale. Purple outlines mark persistent Jev losses, teal persistent gains, and light blue mixed outcomes.}
\label{fig:final-cases}
\end{figure*}

To connect task outcomes with agent behavior, we compare three groups over three runs on fixed tasks, persistent Jev failures with consistent baseline success, the reverse contrast, and mixed outcomes where each configuration succeeds once or twice. They contain nineteen, three, and 21 of the 616 family--task pairs (Figure~\ref{fig:case-summary}), totaling 43 pairs and 258 executions. Figure~\ref{fig:final-cases} places them within the full outcome distribution.

\textbf{Systematic trace coding.} Two authors independently applied seven binary behavior labels to each execution, with disagreements resolved by a third author against the records. Labels can co-occur and include transient errors in successful runs. Substantive action records support coding for 251 executions, including three with partial coverage. Seven executions without action records contribute to outcome counts. Before adjudication, mean per-label Cohen\textquotesingle s $\kappa$ was 0.867, binary agreement was 96.6\%, and exact label-set agreement was 79.7\%. Table~\ref{tab:trace-coding} reports category counts. Appendix~\ref{sec:trace-coding} defines the taxonomy and denominators.

\begin{table}[!htbp]
\centering\footnotesize
\setlength{\tabcolsep}{2pt}
\begin{tabular*}{\columnwidth}{@{\extracolsep{\fill}}lrr@{}}
\toprule
Behavior & Executions (\%) & $\kappa$ \\
\midrule
Target identification & 42 (16.7) & 0.794 \\
Prerequisite violation & 27 (10.8) & 0.781 \\
Failed recovery & 73 (29.1) & 0.882 \\
Premature completion & 14 (5.6) & 0.919 \\
Ineffective search & 62 (24.7) & 0.934 \\
Execution granularity & 2 (0.8) & 1.000 \\
Output or state mismatch & 40 (15.9) & 0.756 \\
\bottomrule
\end{tabular*}
\caption{Behavior labels in 251 codable executions from the outcome-selected set. Counts follow adjudication. Agreement precedes adjudication. Categories overlap.}
\label{tab:trace-coding}
\end{table}

\textbf{Recovery and search distinguish persistent losses.} Among 57 Jev executions in persistent-loss pairs, 40 contain failed recovery and 29 ineffective search, compared with 3 and 2 of 57 baseline executions. FastAgent-J accounts for 39 of the 73 recovery labels, and T3A-J for 11. In mixed pairs, prerequisite violations occur in 13/63 Jev and 6/58 codable baseline executions. These patterns characterize the selected outcome groups and highlight correction, search, and prerequisite handling as recurring difficulties (Appendix~\ref{sec:trace-coding}).

\textbf{Initial success and recovery differ.} Reflexion-J succeeds in 383/402 first trials (95.27\%), exceeding B's 376/402 (93.53\%). It subsequently recovers 11/19 initial failures, compared with 25/26 for B (Table~\ref{tab:reflexion-recovery}). Final success is consequently lower for J. These recovery fractions condition on each configuration's own initial failures. The pattern connects final reliability to sequential correction within the joint selection-and-reflection workflow.

\begin{table}[!htbp]
\centering\footnotesize
\setlength{\tabcolsep}{2pt}
\begin{tabular*}{\columnwidth}{@{\extracolsep{\fill}}lrrr@{}}
\toprule
Reflexion & Initial success & Recovery & Final success \\
\midrule
B-version & 376/402 & 25/26 & 401/402 \\
J-version & 383/402 & 11/19 & 394/402 \\
C-version & 371/402 & 23/31 & 394/402 \\
\bottomrule
\end{tabular*}
\caption{First-trial success, recovery of initial failures, and final success across three runs.}
\label{tab:reflexion-recovery}
\end{table}

\textbf{Reliable selection requires correction and completion checks.} The coded behaviors and Reflexion outcomes connect reliability to how agents respond after an error. Recovery failures sustain unsuccessful execution, while premature completion ends opportunities for correction. These findings support feedback that revises mistaken choices, checks action prerequisites, and verifies the requested final state.

\section{Discussion}

\textbf{Model development.} Our findings point toward decision models trained for the consequences of sequential choices. Jev's fast selection coexists with errors in target interpretation, correction, and termination, and these errors can consume its resource savings through prolonged execution. For Jev and Jev-like models, a promising direction is to retain concise numerical outputs while strengthening sensitivity to goals, action prerequisites, and contradictory feedback. Training on contrasting successful and failed trajectories could target decisions that restore progress or correctly end a task. Another useful objective is calibrated confidence in whether any available action advances the goal, giving the model a basis for requesting more information or reconsidering its choices. Progress should be assessed through successful task delivery and the resources it requires, alongside decision accuracy and response time.

\textbf{Agentic system design.} The engineering implication is to treat decision making as an explicit system component with observable inputs, outcomes, and costs. Candidate interfaces should expose target identity, prerequisites, and expected state changes, with feedback indicating whether the last action advanced the goal. Checks should focus on completion judgments and repeated lack of progress, the points where our traces show failures becoming terminal or persistent. Additional observation or reasoning can then be directed to unresolved decisions. This allocation matters because auxiliary calls and interaction overhead can absorb the savings from faster selection. Deployment decisions should jointly consider the original runtime bottleneck, changes in interaction length, and expenditure per successful outcome. Jev and Jev-like models are consequently most promising in systems that make routine choices inexpensive while concentrating verification and recovery on decisions that threaten completion.

\section{Conclusion}

We evaluate AI agents with Jev-based action selection to examine whether faster, cheaper decisions translate into efficient and reliable task completion. Across five agents over two scenarios, decision savings yield workflow-dependent task gains. Success remains close to baseline on ALFWorld and falls more substantially in two mobile agents. Auxiliary reasoning and additional interactions absorb savings, while failed recovery and premature completion limit delivery. Mean cost per successful outcome rises in one agent despite lower spending per attempt. Agent workflows should preserve progress, support recovery, and verify completion. Efficient decision models should be assessed through successful delivery, recovery behavior, and resources per delivered outcome.

\section*{Limitations}

\textbf{Evaluation scope.} This study examines Jev integration in selected agent families across mobile application and household-task environments. Both settings use language-based observations and bounded action interfaces. Transfer to visually grounded agents, open-ended tool creation, and longer workflows requires further evaluation. The three repetitions characterize execution variability on fixed task instances with a shared seed. They do not measure variation across newly sampled environments or task distributions.

\textbf{System-level comparisons.} Jev operates within an agent workflow whose candidate construction, auxiliary generation, and feedback mechanisms jointly influence performance. Our comparisons therefore concern the evaluated integrations. The ALFWorld LLM-Choice control helps distinguish decision-model effects under a shared candidate interface, but this control does not cover every integration. Within-family budget matching supports comparisons between variants. Comparisons across families also reflect differences in interaction budgets and workflow design. The findings should consequently be interpreted in the context of these configurations, without attributing every observed difference to Jev alone.

\textbf{Measurement conditions.} Latency and cost depend on the evaluated model versions, service conditions, and experimental prices. Hosted inference introduces variation beyond the agent's control, while future model updates or pricing changes may alter the observed trade-offs. Our accounting includes auxiliary model use and failed tasks, but retry adjustments and imputed usage introduce measurement uncertainty where applicable. The reported costs represent estimates under the stated accounting rules, and timing comparisons retain the measurement boundaries of each benchmark.

\bibliography{references}

\appendix
\makeatletter
\setlength{\@fptop}{0pt}
\setlength{\@fpsep}{12pt}
\setlength{\@fpbot}{0pt plus 1fil}
\setlength{\@dblfptop}{0pt}
\setlength{\@dblfpsep}{12pt}
\setlength{\@dblfpbot}{0pt plus 1fil}
\makeatother
\section{Evaluation and Statistical Methods}
\label{sec:final-statistics}

\subsection{Aggregation and Resource Accounting} For run $r$, let $I_r$ contain either all task instances or those with official success, according to the reported scope. A task-level quantity $x_{ir}$ and its action-selection counterpart $a_{ir}$ give
\begin{align}
 m_r^{\mathrm{task}} &= \frac{\sum_{i\in I_r}x_{ir}}{|I_r|}, \\
 m_r^{\mathrm{decision}} &= \frac{\sum_{i\in I_r}a_{ir}}{\sum_{i\in I_r}d_{ir}},
\end{align}
where $d_{ir}$ is the number of decision calls. Each decision has equal weight within a run. For either metric, the reported mean and sample standard deviation are
\begin{align}
 \bar m &= \frac{1}{R}\sum_{r=1}^{R}m_r, \\
 s_m &= \sqrt{\frac{\sum_{r=1}^{R}(m_r-\bar m)^2}{R-1}},
\end{align}
with $R=3$. The full-task scope contains 134 instances per ALFWorld run and 116 per AndroidWorld run.

\textbf{Paired comparisons.} A percentage resource change is $100(\bar m_J/\bar m_L-1)$, where $L$ denotes the specified LLM control. For matched successes, $I_r$ contains the intersection of instances completed by both configurations in run $r$. The task and run identifiers define each pair, and the same formulas apply to both configurations on that subset.

\textbf{Cost accounting.} Prices per million tokens are \$0.20 for LLM input, \$1.20 for LLM output, \$0.042 for Jev input, and zero for Jev output. Task expenditure includes all decision and auxiliary calls. Token counts follow each model's tokenizer. Cost per delivered success divides expenditure over all task--run executions by their total number of successes. Failure expenditure is the fraction of this pooled cost spent on failed executions. Successful-task means instead average resource use only over each configuration's own successful executions.

\subsection{Task-Cluster Bootstrap}
\label{sec:cluster-bootstrap}

For each family and control, we resample $n$ task identities with replacement from its $n$ evaluated tasks, with $n=134$ for ALFWorld and $n=116$ for AndroidWorld. Every sampled task contributes all three runs of both configurations. We compute J minus control differences in success rate, task cost, and task time for 20,000 replicates. The 2.5th and 97.5th percentiles form the 95\% interval. Task resampling describes uncertainty across the evaluated task mix conditional on the fixed instances and run conditions. Computation uses NumPy 1.26.3 with seed 20261006.

For all-task metrics, the estimate is the mean paired difference over tasks and runs. For matched successes, each replicate retains the task--run pairs completed by both configurations and averages the three within-run mean differences. Thus, the matched subset can vary across runs and replicates. For resources per delivered success, each configuration's total cost or time over all attempts is divided by its total successes within each replicate, and the two ratios are differenced. Table~\ref{tab:bootstrap} includes every J--B and J--C comparison. Intervals are pointwise, with no multiplicity adjustment or $p$ values.

\section{Supplementary Experimental Results}
\label{sec:supp-effectiveness}

\subsection{Task Effectiveness}

\textbf{Success, scores, and consistency.} Table~\ref{tab:final-effectiveness} includes 402 executions per ALFWorld configuration and 348 per AndroidWorld configuration. Successes pool all three runs. Score is the AndroidWorld evaluator's mean task score, expressed as a percentage. All-three success counts distinct instances completed in every run. Its denominator is 134 or 116 instances, respectively.

\textbf{Paired outcomes.} Table~\ref{tab:paired-outcomes} retains task--run correspondence. A gain is an execution completed by J but not by the specified control, and a loss is the reverse. Both-success counts determine the sample sizes used for matched resource comparisons. Adding gains to both-success counts gives the number of Jev successes, while adding losses gives the control successes. Their difference is the net change in successful executions. For a given benchmark, dividing this difference by the total executions converts it to the success-rate difference. These counts retain each run separately, so a task can contribute to different outcome categories across repetitions.

\begin{table}[!t]
\centering\fontsize{8}{9.5}\selectfont
\setlength{\tabcolsep}{1.5pt}
\begin{tabular*}{\columnwidth}{@{\extracolsep{\fill}}lrrrr@{}}
\toprule
Configuration & Successes & \shortstack{Success\\rate (\%)} & \shortstack{Score\\(\%)} & \shortstack{All-three\\success} \\
\midrule
\multicolumn{5}{l}{\textbf{ALFWorld}} \\
\agent{ReAct-B} & 392/402 & 97.51 $\pm$ 1.14 & -- & 127/134 \\
\agent{ReAct-J} & 386/402 & 96.02 $\pm$ 0.86 & -- & 124/134 \\
\agent{ReAct-C} & 390/402 & 97.01 $\pm$ 0.75 & -- & 127/134 \\
\agent{Reflexion-B} & 401/402 & 99.75 $\pm$ 0.43 & -- & 133/134 \\
\agent{Reflexion-J} & 394/402 & 98.01 $\pm$ 0.86 & -- & 129/134 \\
\agent{Reflexion-C} & 394/402 & 98.01 $\pm$ 0.43 & -- & 130/134 \\
\midrule
\multicolumn{5}{l}{\textbf{AndroidWorld}} \\
\agent{T3A-B} & 200/348 & 57.47 $\pm$ 4.75 & 57.47 $\pm$ 4.75 & 53/116 \\
\agent{T3A-J} & 154/348 & 44.25 $\pm$ 6.05 & 44.83 $\pm$ 6.08 & 34/116 \\
\agent{AutoDroid-B} & 65/348 & 18.68 $\pm$ 1.99 & 19.68 $\pm$ 1.79 & 18/116 \\
\agent{AutoDroid-J} & 57/348 & 16.38 $\pm$ 1.49 & 17.39 $\pm$ 1.74 & 14/116 \\
\agent{FastAgent-B} & 174/348 & 50.00 $\pm$ 3.76 & 50.14 $\pm$ 4.00 & 38/116 \\
\agent{FastAgent-J} & 111/348 & 31.90 $\pm$ 2.59 & 33.05 $\pm$ 2.60 & 30/116 \\
\bottomrule
\end{tabular*}
\caption{Effectiveness across three runs. Success rates and AndroidWorld official scores are run means $\pm$ sample SD. All-three success counts distinct tasks completed in every run.}
\label{tab:final-effectiveness}
\end{table}
\begin{table}[!t]
\centering\fontsize{8}{9.5}\selectfont
\setlength{\tabcolsep}{1.5pt}
\begin{tabular*}{\columnwidth}{@{\extracolsep{\fill}}llrrr@{}}
\toprule
Family & Control & Gains & Losses & Both \\
\midrule
ReAct & \textit{B} & 8 & 14 & 378 \\
ReAct & \textit{C} & 4 & 8 & 382 \\
Reflexion & \textit{B} & 1 & 8 & 393 \\
Reflexion & \textit{C} & 3 & 3 & 391 \\
T3A & \textit{B} & 9 & 55 & 145 \\
AutoDroid & \textit{B} & 7 & 15 & 50 \\
FastAgent & \textit{B} & 27 & 90 & 84 \\
\bottomrule
\end{tabular*}
\caption{Paired execution outcomes for J versus B or C. Both denotes success by both configurations.}
\label{tab:paired-outcomes}
\end{table}

\subsection{Resource Use and Successful Delivery}
\label{sec:supp-resources}

\textbf{Task and decision resources.} Tables~\ref{tab:resources-alfworld} and~\ref{tab:resources-androidworld} give absolute values for both task scopes. Tokens (k) are thousands of tokens, mUSD is $10^{-3}$ USD, and $\mu$USD is $10^{-6}$ USD. The successful-task group uses each configuration's own successful executions.

\begin{table*}[t]
\centering\footnotesize
\setlength{\tabcolsep}{4pt}
\begin{tabular*}{\textwidth}{@{\extracolsep{\fill}}lrrrrrr@{}}
\toprule
Configuration & \multicolumn{3}{c}{Per task} & \multicolumn{3}{c}{Per decision} \\
\cmidrule(lr){2-4}\cmidrule(lr){5-7}
 & Tokens (k) & Cost (mUSD) & Time (s) & Tokens (k) & Cost ($\mu$USD) & Time (s) \\
\midrule
\multicolumn{7}{l}{\textbf{All tasks}} \\
\agent{ReAct-B} & 33.73 $\pm$ 1.15 & 7.74 $\pm$ 0.30 & 71.17 $\pm$ 7.50 & 2.04 $\pm$ 0.07 & 468 $\pm$ 15 & 4.13 $\pm$ 0.49 \\
\agent{ReAct-J} & 58.23 $\pm$ 2.45 & 3.80 $\pm$ 0.26 & 37.33 $\pm$ 5.00 & 3.15 $\pm$ 0.02 & 122 $\pm$ 1 & 1.08 $\pm$ 0.21 \\
\agent{ReAct-C} & 37.63 $\pm$ 2.53 & 8.75 $\pm$ 0.66 & 86.02 $\pm$ 23.49 & 2.51 $\pm$ 0.04 & 582 $\pm$ 15 & 5.68 $\pm$ 1.46 \\
\agent{Reflexion-B} & 38.86 $\pm$ 3.25 & 8.98 $\pm$ 0.67 & 99.55 $\pm$ 20.22 & 2.08 $\pm$ 0.06 & 479 $\pm$ 9 & 5.14 $\pm$ 0.81 \\
\agent{Reflexion-J} & 77.36 $\pm$ 13.83 & 6.05 $\pm$ 1.70 & 61.90 $\pm$ 8.66 & 3.19 $\pm$ 0.07 & 124 $\pm$ 3 & 1.04 $\pm$ 0.05 \\
\agent{Reflexion-C} & 68.35 $\pm$ 0.89 & 16.75 $\pm$ 0.15 & 143.03 $\pm$ 18.64 & 2.71 $\pm$ 0.01 & 670 $\pm$ 1 & 5.84 $\pm$ 0.71 \\
\midrule
\multicolumn{7}{l}{\textbf{Successful tasks}} \\
\agent{ReAct-B} & 31.64 $\pm$ 0.94 & 7.24 $\pm$ 0.20 & 68.18 $\pm$ 8.56 & 2.01 $\pm$ 0.07 & 461 $\pm$ 15 & 4.16 $\pm$ 0.56 \\
\agent{ReAct-J} & 48.64 $\pm$ 1.82 & 2.59 $\pm$ 0.09 & 28.36 $\pm$ 4.35 & 3.10 $\pm$ 0.01 & 119 $\pm$ 1 & 1.07 $\pm$ 0.21 \\
\agent{ReAct-C} & 31.93 $\pm$ 1.33 & 7.26 $\pm$ 0.36 & 76.14 $\pm$ 20.75 & 2.44 $\pm$ 0.03 & 553 $\pm$ 9 & 5.64 $\pm$ 1.52 \\
\agent{Reflexion-B} & 37.57 $\pm$ 1.73 & 8.69 $\pm$ 0.32 & 96.95 $\pm$ 15.73 & 2.06 $\pm$ 0.05 & 476 $\pm$ 7 & 5.16 $\pm$ 0.84 \\
\agent{Reflexion-J} & 51.37 $\pm$ 1.89 & 2.88 $\pm$ 0.32 & 36.60 $\pm$ 4.32 & 3.10 $\pm$ 0.01 & 119 $\pm$ 1 & 1.04 $\pm$ 0.05 \\
\agent{Reflexion-C} & 39.36 $\pm$ 7.54 & 9.15 $\pm$ 1.95 & 87.82 $\pm$ 25.38 & 2.49 $\pm$ 0.09 & 575 $\pm$ 32 & 5.34 $\pm$ 0.72 \\
\bottomrule
\end{tabular*}
\caption{ALFWorld resource metrics. Each entry is a mean $\pm$ sample SD across three runs. The successful-task group includes only successful executions and their decisions. Decision averages divide action-selection totals by decision calls within each run. Cost units are $10^{-3}$ USD per task and $10^{-6}$ USD per decision.}
\label{tab:resources-alfworld}
\end{table*}

\begin{table*}[t]
\centering\footnotesize
\setlength{\tabcolsep}{4pt}
\begin{tabular*}{\textwidth}{@{\extracolsep{\fill}}lrrrrrr@{}}
\toprule
Configuration & \multicolumn{3}{c}{Per task} & \multicolumn{3}{c}{Per decision} \\
\cmidrule(lr){2-4}\cmidrule(lr){5-7}
 & Tokens (k) & Cost (mUSD) & Time (s) & Tokens (k) & Cost ($\mu$USD) & Time (s) \\
\midrule
\multicolumn{7}{l}{\textbf{All tasks}} \\
\agent{T3A-B} & 187.76 $\pm$ 20.02 & 39.48 $\pm$ 3.95 & 213.00 $\pm$ 4.19 & 6.55 $\pm$ 0.73 & 1413 $\pm$ 147 & 7.24 $\pm$ 0.35 \\
\agent{T3A-J} & 288.44 $\pm$ 58.30 & 32.73 $\pm$ 6.81 & 190.36 $\pm$ 27.06 & 12.86 $\pm$ 1.54 & 480 $\pm$ 58 & 1.34 $\pm$ 0.41 \\
\agent{AutoDroid-B} & 12.64 $\pm$ 0.72 & 5.22 $\pm$ 0.38 & 261.20 $\pm$ 8.02 & 1.06 $\pm$ 0.05 & 436 $\pm$ 11 & 8.13 $\pm$ 1.03 \\
\agent{AutoDroid-J} & 34.30 $\pm$ 0.28 & 1.63 $\pm$ 0.04 & 246.27 $\pm$ 10.82 & 2.48 $\pm$ 0.02 & 94 $\pm$ 1 & 1.75 $\pm$ 0.47 \\
\agent{FastAgent-B} & 47.68 $\pm$ 8.80 & 10.98 $\pm$ 1.96 & 119.68 $\pm$ 16.38 & 4.78 $\pm$ 0.02 & 1102 $\pm$ 10 & 5.40 $\pm$ 0.27 \\
\agent{FastAgent-J} & 144.11 $\pm$ 16.06 & 6.26 $\pm$ 0.62 & 120.50 $\pm$ 10.66 & 9.72 $\pm$ 0.29 & 346 $\pm$ 11 & 1.41 $\pm$ 0.23 \\
\midrule
\multicolumn{7}{l}{\textbf{Successful tasks}} \\
\agent{T3A-B} & 149.79 $\pm$ 31.95 & 31.24 $\pm$ 6.46 & 162.00 $\pm$ 6.59 & 6.35 $\pm$ 0.78 & 1346 $\pm$ 155 & 6.12 $\pm$ 0.57 \\
\agent{T3A-J} & 191.85 $\pm$ 25.14 & 21.50 $\pm$ 2.95 & 136.20 $\pm$ 14.46 & 12.23 $\pm$ 1.23 & 455 $\pm$ 46 & 1.37 $\pm$ 0.49 \\
\agent{AutoDroid-B} & 5.23 $\pm$ 0.76 & 1.98 $\pm$ 0.35 & 145.05 $\pm$ 15.81 & 0.91 $\pm$ 0.02 & 343 $\pm$ 8 & 7.44 $\pm$ 1.19 \\
\agent{AutoDroid-J} & 12.27 $\pm$ 1.07 & 0.52 $\pm$ 0.07 & 126.24 $\pm$ 13.06 & 2.02 $\pm$ 0.06 & 76 $\pm$ 2 & 1.47 $\pm$ 0.90 \\
\agent{FastAgent-B} & 42.32 $\pm$ 2.92 & 9.65 $\pm$ 0.60 & 104.89 $\pm$ 6.42 & 4.61 $\pm$ 0.09 & 1053 $\pm$ 19 & 4.47 $\pm$ 0.45 \\
\agent{FastAgent-J} & 60.72 $\pm$ 1.32 & 3.39 $\pm$ 0.02 & 86.21 $\pm$ 7.32 & 7.60 $\pm$ 0.26 & 265 $\pm$ 9 & 1.33 $\pm$ 0.21 \\
\bottomrule
\end{tabular*}
\caption{AndroidWorld resource metrics. Each entry is a mean $\pm$ sample SD across three runs. The successful-task group includes only successful executions and their decisions. Decision averages divide action-selection totals by decision calls within each run. Cost units are $10^{-3}$ USD per task and $10^{-6}$ USD per decision.}
\label{tab:resources-androidworld}
\end{table*}

\textbf{Time components.} Table~\ref{tab:absolute-time} provides component means in seconds per task, pooled over three equally sized runs. Decision and auxiliary columns sum their respective call durations. Environment interaction and other overhead is the residual after subtracting both call components from total task time. Auxiliary cost is its share of pooled inference expenditure.

\begin{table}[!htbp]
\centering\footnotesize
\setlength{\tabcolsep}{2pt}
\begin{tabular*}{\columnwidth}{@{\extracolsep{\fill}}lrrrr@{}}
\toprule
Configuration & \shortstack{Decision\\(s)} & \shortstack{Auxiliary\\(s)} & \shortstack{Env. \&\\other (s)} & \shortstack{Aux. cost\\(\%)} \\
\midrule
\agent{ReAct-B} & 68.31 & 0.00 & 2.86 & 0.0 \\
\agent{ReAct-J} & 17.25 & 17.23 & 2.85 & 48.9 \\
\agent{ReAct-C} & 79.18 & 3.95 & 2.89 & 7.7 \\
\agent{Reflexion-B} & 95.76 & 0.89 & 2.90 & 1.0 \\
\agent{Reflexion-J} & 20.16 & 38.57 & 3.17 & 60.3 \\
\agent{Reflexion-C} & 119.76 & 20.17 & 3.10 & 18.0 \\
\agent{T3A-B} & 88.52 & 67.47 & 57.01 & 56.3 \\
\agent{T3A-J} & 16.14 & 96.99 & 77.24 & 81.9 \\
\agent{AutoDroid-B} & 96.78 & 0.00 & 164.43 & 0.0 \\
\agent{AutoDroid-J} & 23.34 & 9.56 & 213.37 & 22.9 \\
\agent{FastAgent-B} & 52.99 & 2.27 & 64.42 & 1.0 \\
\agent{FastAgent-J} & 20.03 & 14.76 & 85.71 & 21.1 \\
\bottomrule
\end{tabular*}
\caption{Absolute task-time components and auxiliary inference expenditure over all executions.}
\label{tab:absolute-time}
\end{table}

\textbf{Successful delivery.} Table~\ref{tab:matched-delivery} reports resource changes on common successful task--run pairs, alongside failure expenditure and cost per delivered success. Failure expenditure is the fraction of total inference cost consumed by failed attempts. Cost per delivered success divides inference cost across all attempts by the number of successes.

\subsection{Paired Effect Estimates}
\label{sec:paired-estimates}

Table~\ref{tab:bootstrap} reports J minus control differences for all executions, matched successes, and resources per delivered success. Intervals use 20,000 paired task-cluster bootstrap replicates, retaining all three runs of both configurations for each sampled task. Brackets show the 2.5th and 97.5th percentiles.

\begin{table*}[t]
\centering\footnotesize
\setlength{\tabcolsep}{3pt}
\begin{tabular*}{\textwidth}{@{\extracolsep{\fill}}llrrr@{}}
\toprule
Family & Control & $\Delta$ Success (pp) & $\Delta$ Cost (mUSD) & $\Delta$ Time (s) \\
\midrule
\multicolumn{5}{l}{\textbf{All task executions}} \\
ReAct & B & -1.49 [-4.48, +1.24] & -3.94 [-4.88, -2.94] & -33.84 [-42.51, -24.89] \\
ReAct & C & -1.00 [-3.23, +0.75] & -4.95 [-6.11, -3.84] & -48.69 [-61.13, -37.16] \\
Reflexion & B & -1.74 [-3.98, +0.00] & -2.93 [-6.05, +0.90] & -37.65 [-61.89, -9.55] \\
Reflexion & C & +0.00 [-1.00, +1.00] & -10.70 [-16.90, -5.93] & -81.13 [-132.09, -42.03] \\
T3A & B & -13.22 [-17.82, -8.62] & -6.75 [-10.05, -3.44] & -22.64 [-42.08, -3.72] \\
AutoDroid & B & -2.30 [-5.46, +0.86] & -3.59 [-4.08, -3.11] & -14.94 [-32.63, +3.02] \\
FastAgent & B & -18.10 [-26.44, -9.77] & -4.72 [-6.36, -3.09] & +0.82 [-9.63, +12.77] \\
\midrule
\multicolumn{5}{l}{\textbf{Matched successful executions}} \\
ReAct & B & -- & -4.68 [-5.24, -4.15] & -40.55 [-47.70, -33.90] \\
ReAct & C & -- & -4.56 [-5.34, -3.86] & -47.85 [-58.95, -37.98] \\
Reflexion & B & -- & -5.61 [-6.98, -4.46] & -56.38 [-70.44, -44.23] \\
Reflexion & C & -- & -5.44 [-6.82, -4.32] & -38.31 [-51.38, -26.92] \\
T3A & B & -- & -2.64 [-4.71, -0.39] & +1.49 [-12.98, +17.24] \\
AutoDroid & B & -- & -1.48 [-2.06, -1.01] & -26.88 [-42.70, -13.86] \\
FastAgent & B & -- & -2.72 [-3.39, -2.14] & -4.49 [-9.87, +1.32] \\
\midrule
\multicolumn{5}{l}{\textbf{Resources per delivered success}} \\
ReAct & B & -- & -3.98 [-5.07, -2.81] & -34.11 [-43.92, -23.82] \\
ReAct & C & -- & -5.06 [-6.41, -3.81] & -49.79 [-63.41, -37.03] \\
Reflexion & B & -- & -2.83 [-6.06, +1.30] & -36.64 [-61.92, -6.47] \\
Reflexion & C & -- & -10.91 [-17.70, -5.93] & -82.77 [-138.08, -42.02] \\
T3A & B & -- & +5.27 [-3.83, +17.34] & +59.55 [+7.44, +126.66] \\
AutoDroid & B & -- & -18.00 [-29.85, -11.68] & +105.07 [-206.65, +559.67] \\
FastAgent & B & -- & -2.33 [-7.60, +4.38] & +138.43 [+60.33, +257.46] \\
\bottomrule
\end{tabular*}
\caption{Paired differences for J minus the specified control with pointwise 95\% task-cluster bootstrap intervals. All three runs of each sampled task remain together. Matched-success estimates average the three run-specific paired means. Delivered-success estimates divide total resources across all attempts by total successes.}
\label{tab:bootstrap}
\end{table*}

\section{Trace Analysis and Case Evidence}
\subsection{Coding Protocol and Behavioral Patterns}
\label{sec:trace-coding}

The unit of annotation is one execution. The selected set contains all 43 family--task pairs in the persistent-loss, persistent-gain, and mixed-outcome groups defined in RQ3. Each contributes three baseline and three Jev runs, giving 258 executions. Selection uses official outcomes, so category frequencies describe these groups. Two authors independently applied the same codebook to the goals, outcomes, and action records without access to each other's annotations. A third author resolved disagreements using actions, observations, and final UI states. The annotators consulted the original records when the extracted records were insufficient. Agreement was calculated from the two independent annotations before adjudication. The independent annotations contained 51 label-set disagreements and one coverage disagreement.

\textbf{Behavior definitions.} Target identification marks a wrong object, entity, or UI target. Prerequisite violation marks an attempted action or search with an unmet state or inventory condition. Failed recovery marks repeated unsuccessful operations or persistence in a contradicted premise after feedback. Premature completion marks an explicit completion claim while a goal condition remains unmet. Ineffective search marks repeated unproductive search, navigation, or strategy changes. Execution granularity marks an observed error from batched operations or stale action references. Output or state mismatch marks an explicit discrepancy between the delivered answer or state and the requested result or evaluator criterion. A failed outcome alone provides no behavior label. An execution can receive several labels, and successful executions can contain corrected errors.

\textbf{Coverage and agreement.} The behavioral denominator includes 248 complete and three partial records. The partial records contain T3A summary-call errors but retain substantive actions. Six FastAgent baseline executions contain no action events, and one contains narrative responses without tool execution. These seven executions are excluded from behavior counts and agreement. For each label, Cohen's $\kappa=(p_o-p_e)/(1-p_e)$ compares the two original binary annotations, with $p_e$ computed from their marginal positive rates. The reported mean is the unweighted mean over seven labels. Binary agreement pools the 1,757 execution--label decisions, and exact-set agreement requires agreement on all seven labels for an execution. Table~\ref{tab:trace-coding} gives individual $\kappa$ values. The granularity label has two positives for each annotator.

\textbf{Behavioral patterns.} Table~\ref{tab:coding-groups} reports adjudicated counts by agent and outcome group. Each denominator includes executions with substantive action records. Labels can overlap, and the counts describe the selected outcome groups.

\begin{table*}[t]
\centering\footnotesize
\setlength{\tabcolsep}{3pt}
\begin{tabular*}{\textwidth}{@{\extracolsep{\fill}}llrrrrrrrr@{}}
\toprule
Group & Variant & $n$ & Target & Prereq. & Recovery & Premature & Search & Granularity & Output \\
\midrule
\multicolumn{10}{l}{\textbf{Family}} \\
AutoDroid & B & 15 & 0 & 0 & 1 & 0 & 2 & 0 & 5 \\
AutoDroid & J & 15 & 0 & 0 & 6 & 0 & 3 & 0 & 4 \\
FastAgent & B & 62 & 5 & 4 & 1 & 1 & 4 & 2 & 3 \\
FastAgent & J & 69 & 10 & 7 & 39 & 5 & 32 & 0 & 9 \\
ReAct & B & 9 & 1 & 7 & 3 & 0 & 3 & 0 & 0 \\
ReAct & J & 9 & 6 & 0 & 0 & 3 & 0 & 0 & 3 \\
Reflexion & B & 3 & 1 & 1 & 1 & 0 & 1 & 0 & 0 \\
Reflexion & J & 3 & 3 & 0 & 3 & 0 & 0 & 0 & 0 \\
T3A & B & 33 & 9 & 2 & 8 & 2 & 6 & 0 & 7 \\
T3A & J & 33 & 7 & 6 & 11 & 3 & 11 & 0 & 9 \\
\midrule
\multicolumn{10}{l}{\textbf{Outcome group}} \\
Persistent Jev loss & B & 57 & 5 & 8 & 3 & 0 & 2 & 1 & 0 \\
Persistent Jev loss & J & 57 & 20 & 0 & 40 & 8 & 29 & 0 & 14 \\
Persistent Jev gain & B & 7 & 1 & 0 & 0 & 0 & 3 & 1 & 2 \\
Persistent Jev gain & J & 9 & 0 & 0 & 0 & 0 & 0 & 0 & 0 \\
Mixed outcomes & B & 58 & 10 & 6 & 11 & 3 & 11 & 0 & 13 \\
Mixed outcomes & J & 63 & 6 & 13 & 19 & 3 & 17 & 0 & 11 \\
\bottomrule
\end{tabular*}
\caption{Adjudicated behavior counts by family and selected outcome group. $n$ counts executions with substantive action records. All columns count executions bearing the label and can overlap.}
\label{tab:coding-groups}
\end{table*}

\subsection{Case Evidence and Execution Records}
\label{sec:final-trajectories}

\textbf{Task identities and coverage.} Table~\ref{tab:case-records} identifies the illustrative tasks and preserves the order of their three run outcomes. ALFWorld identifiers refer to the \texttt{valid\_unseen} split. AndroidWorld names identify task types, each evaluated on one fixed parameter instance. The full case grouping comprises 268 ALFWorld and 348 AndroidWorld family--task pairs. Each selected pair contributes six executions, three per configuration. A code of 111 records success in every run and 000 records failure in every run. Codes such as 010 and 101 retain the positions of variable outcomes. Thus, the two configurations in ALFWorld~133 succeed in complementary runs, whereas both configurations in the recipe-deletion example succeed in runs~1 and~3. The table records task completion, and the action descriptions in this appendix specify what happened within the corresponding executions.
\textbf{Record coverage.} The recipe-creation field-shift account uses run~1, where the baseline batches operations against a changing form. Its other two failed runs contribute to the task's outcome group. Behavioral descriptions use substantive action records, while all group counts use official task outcomes.

\begin{table}[!htbp]
\centering\footnotesize
\setlength{\tabcolsep}{2pt}
\begin{tabular*}{\columnwidth}{@{\extracolsep{\fill}}lp{3.05cm}rr@{}}
\toprule
Family & Task identifier & B & J \\
\midrule
ReAct & 046 & 111 & 000 \\
ReAct & 101 & 111 & 000 \\
Reflexion & 045 & 111 & 000 \\
ReAct & 133 & 010 & 101 \\
FastAgent & SystemWifiTurnOn & 111 & 000 \\
FastAgent & RecipeAddMultiple Recipes & 000 & 111 \\
FastAgent & SportsTrackerTotal DurationForCategory ThisWeek & 000 & 111 \\
FastAgent & OsmAndFavorite & 000 & 111 \\
FastAgent & RecipeDeleteMultiple RecipesWithNoise & 101 & 101 \\
\bottomrule
\end{tabular*}
\caption{Official outcomes in runs 1, 2, and 3. Each digit is success (1) or failure (0). The first four rows are ALFWorld and the others AndroidWorld.}
\label{tab:case-records}
\end{table}

\textbf{Persistent Jev failures.}\nopagebreak[4]
\label{sec:cases-losses}

Of the nineteen pairs, three occur in ALFWorld and sixteen in AndroidWorld, including fourteen for FastAgent. Household cases concern cleaning or heating, while mobile cases span settings, records, calendars, and media. Representative failures involve persisting with an incorrect target or accepting an unmet goal at termination.

In cloth cleaning (ALFWorld~046), \agent{ReAct-J} selects a handtowel in all three runs and repeatedly invokes THINK when cleaning remains unavailable. The baseline makes the same initial error in run~2, but puts down the handtowel, retrieves a cloth, and completes the task. Correction distinguishes these outcomes. In Reflexion task~045, auxiliary LLM reflections preserve the wrong-object premise across ten unsuccessful trials in run~1. Additional reasoning reinforces the mistake. In the Wi-Fi task, \agent{FastAgent-J} opens quick settings, clicks a switch labeled ``Off'', and declares success after three decisions in all three runs, while Wi-Fi remains off. The label provides no Wi-Fi identity. The object cases sustain an initial error through failed correction, while the Wi-Fi case terminates before the required state is reached.

\textbf{Persistent Jev gains.}\nopagebreak[4]
\label{sec:cases-gains}

All three reverse pairs belong to FastAgent, involving recipe creation, activity-duration reporting, and adding a map favorite. They expose distinct contributions from interaction granularity, response format, and the evaluator's acceptance criterion. This reverse contrast connects goal-appropriate choices with current-state execution and delivery in the requested response format.

In recipe creation, the first baseline run batches field operations and reports that a value entered the wrong field after the form shifted. Jev completes the task through single-tool steps with updated observations between operations, keeping choices grounded in the current layout. For the activity query, both workflows obtain 105 minutes in run~1. Jev returns the requested integer ``105'', while the baseline returns a sentence containing that value and fails the evaluator. Correct calculation and compliant output are separate requirements. In the map task, the first baseline run expands its search and stops when it cannot verify the location. Jev creates the requested named favorite and passes the name-based evaluator. This contrast separates geographic verification from benchmark acceptance. Successful delivery depends on execution, response format, and completion criteria.

\textbf{Mixed outcomes.}\nopagebreak[4]
\label{sec:cases-mixed}

The 21 mixed pairs comprise one ReAct, ten T3A, four AutoDroid, and six FastAgent tasks. Twenty occur in AndroidWorld. Fixed task instances make these contrasts informative about variation within the same workflow. Representative runs differ in prerequisite management and search strategy, showing inconsistent application of effective plans.

In two-object placement (ALFWorld~133), \agent{ReAct-J} succeeds twice and \agent{ReAct-B} once. The failed Jev run holds the first object while repeatedly reasoning about collecting the second, whose inventory prerequisite is unmet. Successful runs deposit the first object before collecting the next. Progress requires making a needed action available as well as choosing among currently admissible actions. In recipe deletion among irrelevant entries, both FastAgent variants succeed twice. The failed Jev run deletes two targets and scrolls for the third until its budget expires. Successful Jev runs locate the targets through search, showing that strategy choice varies within the same configuration.

\textbf{Action positions.} In ALFWorld~046, \agent{ReAct-B} run~2 takes the handtowel at step~3. Cleaning attempts at steps~5 and~12 return ``Nothing happens''. It deposits the handtowel at step~28, retrieves cloth~2 at step~32, cleans it at step~34, and places it in the cabinet at step~36. In the additional cup-heating example, ALFWorld~101, \agent{ReAct-J} run~1 refers to mug~2 as the requested cup and continues making completion assertions through step~49. In recipe deletion with irrelevant entries, \agent{FastAgent-J} run~2 performs 24 swipes after deleting two targets and ends with the third target still present.

\textbf{Completion fields and evaluator outcomes.} For the activity-duration query in run~1, the Jev workflow's completion call carries the message ``105'' with its internal success field set to false. The returned answer is ``105'' and receives official success. For the map-favorite instance, the evaluator accepts the stored favorite by its requested name. The geographic verification attempted by the baseline is separate from this acceptance condition.

\textbf{Feedback and completion checks.} Across these cases, reliability hinges on observations changing consequential choices. Object mismatches require revised targets, changing forms require updated action references, and stalled progress calls for another strategy. Reflection that preserves a wrong premise adds little corrective value, while accepting an unverified goal ends execution prematurely. Together with the distinction between real-world verification and benchmark acceptance, these findings favor checks tied to target identity, prerequisites, and the requested final state.

\end{document}